\def\year{2022}\relax
\documentclass[letterpaper]{article} 
\usepackage{aaai22}  
\usepackage{times}  
\usepackage{helvet}  
\usepackage{courier}  
\usepackage[hyphens]{url}  
\usepackage{graphicx} 
\usepackage{natbib}  
\usepackage{caption} 
\DeclareCaptionStyle{ruled}{labelfont=normalfont,labelsep=colon,strut=off} 
\usepackage{algorithm}
\usepackage{algorithmic}

\usepackage{multirow}
\usepackage{booktabs} 
\usepackage{bbding}  

\usepackage{amsthm,amsmath,amssymb}  
\usepackage{mathrsfs}
\usepackage{makecell} 

\usepackage{newfloat}
\usepackage{listings}
\floatstyle{ruled}
\newfloat{listing}{tb}{lst}{}
\floatname{listing}{Listing}
\title{Weakly-Supervised Salient Object Detection Using Point Supervision}

\author{
	Shuyong Gao\textsuperscript{\rm 1}, 
	Wei Zhang\textsuperscript{\rm 1*}, Yan Wang\textsuperscript{\rm 1}, Qianyu Guo\textsuperscript{\rm 1}
	\\
	Chenglong Zhang\textsuperscript{\rm 1}, Yangji He\textsuperscript{\rm 1}, Wenqiang Zhang\textsuperscript{\rm 1,2\footnote{Corresponding author}} \\ 
}

\affiliations{
	\textsuperscript{\rm 1}Shanghai Key Laboratory of Intelligent Information Processing, School of Computer Science, Fudan University \\ 
	
	\textsuperscript{\rm 2}Academy for Engineering \& Technology, Fudan University \\
	\{sygao18,weizh,yanwang19,qyguo20,clzhang20,yjhe20,wqzhang\}@fudan.edu.cn
	
}

\usepackage{bibentry}

\begin{document}
	
	\maketitle
	
	\begin{abstract}
		
		Current state-of-the-art saliency detection models rely heavily on large datasets of accurate pixel-wise annotations, but manually labeling pixels is time-consuming and labor-intensive. There are some weakly supervised methods developed for alleviating the problem, such as image label, bounding box label, and scribble label, while point label still has not been explored in this field. In this paper, we propose a novel weakly-supervised salient object detection method using point supervision. To infer the saliency map, we first design an adaptive masked flood filling algorithm to generate pseudo labels. Then we develop a transformer-based point-supervised saliency detection model to produce the first round of saliency maps. However, due to the sparseness of the label, the weakly supervised model tends to degenerate into a general foreground detection model. To address this issue, we propose a Non-Salient Suppression (NSS) method to optimize the erroneous saliency maps generated in the first round and leverage them for the second round of training. Moreover, we build a new point-supervised dataset (P-DUTS) by relabeling the DUTS dataset. In P-DUTS, there is only one labeled point for each salient object. Comprehensive experiments on five largest benchmark datasets demonstrate our method outperforms the previous state-of-the-art methods trained with the stronger supervision and even surpass several fully supervised state-of-the-art models. The code is available at: https://github.com/shuyonggao/PSOD.

	\end{abstract}
	
	\section{Introduction}
	
	\begin{figure}[t]
		\centering
		\includegraphics[width=0.7\columnwidth]{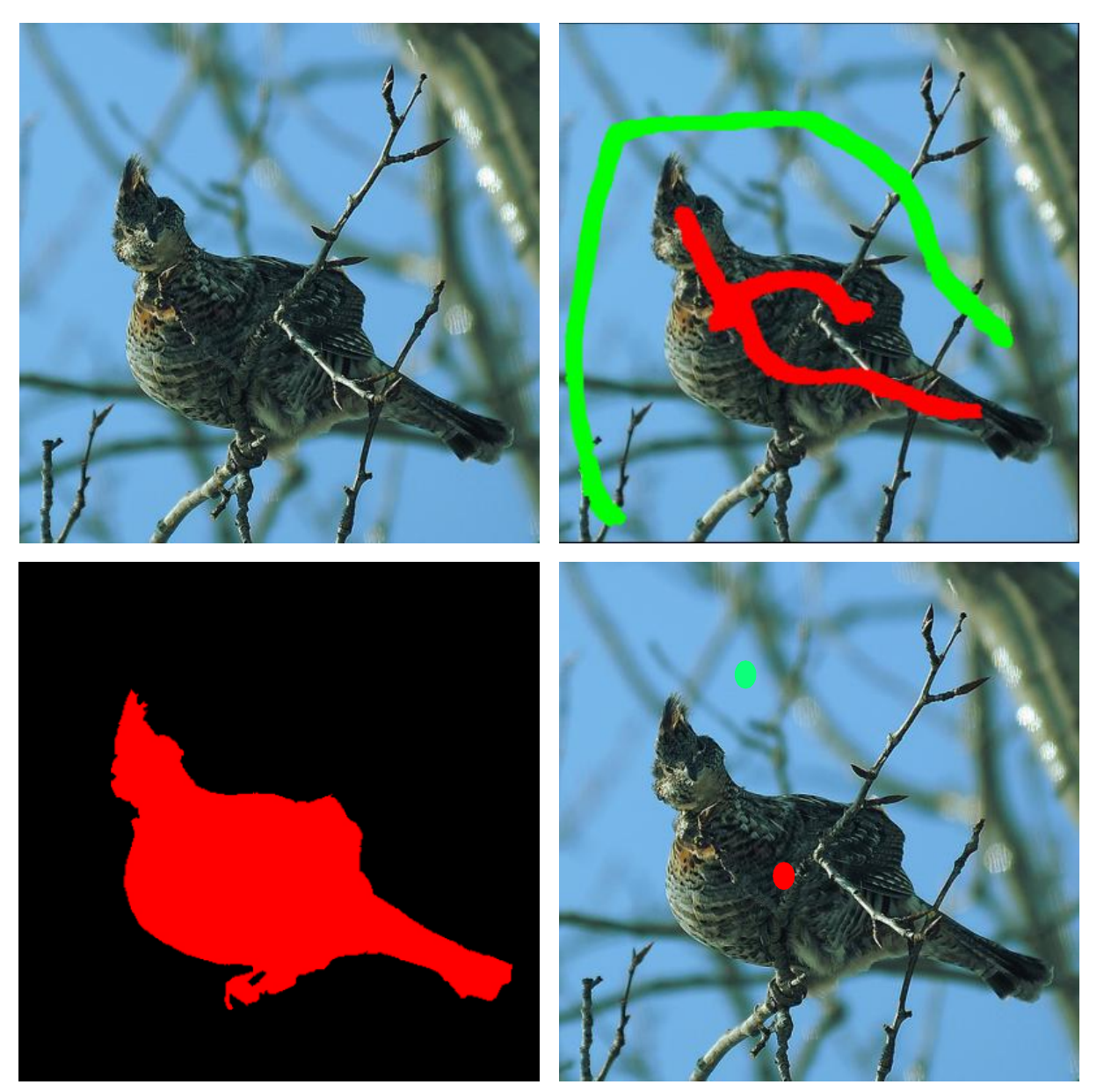} 
		\caption{Top left: original image. Bottom left: fully supervised annotation. Top right: scribble annoation. Bottom right: point annotation.}
		\label{annotation}
	\end{figure}
	
	Salient object detection (SOD), aiming at detecting the most attractive regions of the images or videos to imitate human attention mechanism \cite{depth-survey}, has currently caught increasing attention in image processing, which can be further used in various computer vision fields, such as object recognition \cite{saliency-recognition}, image caption \cite{re-caption}, person re-identification \cite{person-re-id}. Fully supervised salient object detection models have achieved superior performance relying on effectively designed modules \cite{ldf, f3net}. However, these methods heavily rely on large datasets of precise pixel-wise annotations which is a tedious and inefficient procedure. 
	
	Recently, sparse labeling methods have attracted increasing attention, which aims to achieve a trade-off between time consumption and performance. 
	Some methods attempt to use image-level labels to detect salient objects \cite{wsi, duts, mfnet}. And some approaches train the network using noisy labels which are often generated from other SOD methods \cite{tipweaklyboundingbox}. Scribble annotation \cite{wssa} is proposed by lower labeling time consumption. Besides, as compared to previous methods, it can provide local ground truth labels. 
	\cite{wssa} claims that they can finish annotating an image in 1$\sim$2 seconds, but we find it difficult to achieve this for untrained annotators. We attempt to propose a point supervised saliency detection method to achieve an acceptable performance using a less time-consuming method compared to scribble annotation. As far as we know, this is the first attempt to accomplish salient object detection using point supervision (Fig. \ref{annotation}).

	In this paper, we propose a framework that uses point supervision to train the salient object model. We build a new weakly supervised saliency detection annotation dataset by relabeling DUTS \cite{duts} dataset using point supervision.   We choose to use point annotation to label the dataset for the following reasons: Firstly, point supervision can provide relatively accurate location information directly compared to unsupervised and image-level supervision. Secondly, point supervision is probably the least time-costly annotation method compared to other manual annotation methods. 
	
	Moreover, we find that the current weakly supervised saliency model \cite{wssa, scwssod, msw} only focuses on objects which must be noticed in the corresponding scenario, but overlooks objects that should be ignored. The reason is that due to the sparsity of weakly supervised labels, the supervision signal can only cover a small area of the images, which simply allows the model to learn which objects must be highlighted, but lacks information to guide the model on which ones should be ignored. As described at the beginning, the SOD is to simulate human visual attention and filter out non-salient objects. In a scene, the latter is the core of attention, as segmenting all objects means that "human attention" is not formed. At this point, the model degenerates into a model that segments all the objects that have been seen. Based on this observation, we propose the Non-Salient Suppression (NSS) method to explicitly filter out the non-salient but detected objects.
	
	As weakly supervised labels only cover part of the object regions which makes it difficult for the model to perceive the structure of the objects, \cite{wssa} leverage an edge detector \cite{edge_richer} to generate edges to supervise the training process, thus indirectly complementing the structural cues. Our method to generate initial pseudo-labels is simple yet very effective. We observe that the detected edges not only provide structure to the model but also divide the image into multiple connected regions. Inspired by the flood filling algorithm (a classical image processing algorithm), the point annotations and the edges are exploited to generate the initial pseudo label as shown in Fig. 2.
	Due to the fact that the edges are frequently discontinuous and blurred, we design an adaptive mask to address this issue. In summary, our main contributions can be summarized as follows:

	\begin{itemize}
		\item We propose a novel weakly-supervised salient object detection framework to detect salient objects by single point annotation, and build a new point-supervised saliency dataset P-DUTS.
		
		\item We uncover the problem of degradation of weakly supervised saliency detection models and propose the Non-Salient object Suppression (NSS) method to explicitly filter out the non-salient but detected objects.
		
		\item We design a transform-based point supervised SOD model that, in collaboration with the proposed adaptive flood filling, not only outperforms existing state-of-the-art weakly-supervised methods with stronger supervision by a large margin, even surpasses many fully-supervised methods. 
	\end{itemize}

	\section{Related Work}

	\subsection{Weakly Supervised Saliency Detection}
	With the development of weakly supervised learning in dense prediction tasks, some works attempt to employ sparse annotation to acquire the salient objects. \cite{sbf} leverages the fused saliency maps provided by several unsupervised heuristic SOD methods as pseudo labels to train the final saliency model. By incorporating dense Conditional Random Field (CRF) \cite{crf} as post-processing steps, \cite{wsi,wssa} further optimized the resulting results. \cite{msw} use multimodal data, including category labels, captions, and noisy pseudo labels to train the saliency model where class activation maps (CAM) \cite{cam} is used as the initial saliency maps. Recently, scribble annotation for saliency detection, a more user-friendly annotation method, is proposed by \cite{wssa}, in which edge generated by edge detector \cite{edge_richer} is used to furnish structural information. Further, \cite{scwssod} enhances the structural consistency of the saliency maps by introducing saliency structure consistency loss and gated CRF loss \cite{gate_crf} to improve the performance of the model on scribble dataset \cite{wssa}.
	
	\subsection{Point Annotation in Similar Field}
	There has been several works on point annotations in weakly supervised segmentation \cite{what_point, metric_point} and instance segmentation \cite{point_large, point_latent, point_extreme, point_regional}.
	Semantic segmentation focuses on class-specific categories, but saliency detection does not focus on category properties,
	and it often occurs that an object is significant in one scenario, but not in another. That is, saliency detection is dedicated to the contrast between the object and the current scene \cite{itti}. Point-based instance segmentation methods are mainly used in interactive pipelines.

	\subsection{Vision Transformer} 
	Visual processing-oriented transformer introduced from Transformer in Natural Language Processing (NLP) has attracted much attention in multiple fields of computer vision. Vision Transformer (ViT) \cite{vit} introduces a pure transformer model in computer vision tasks and achieves state-of-the-art performance on the image classification task. By employing ViT as the backbone, Segmentation Transformer (SETR) \cite{setr} add a decoder head to generate the final segmentation results with minimal modification. Detection Transformer \cite{detr} leverage encoder-decoder transformer and standard convolution network on object detection. Considering the scale difference between natural language and images, \cite{swin} design a shifted window scheme to improve efficiency as well as accuracy.

	\begin{figure*}[htbp]
		\centering
		\includegraphics[width=0.75\textwidth]{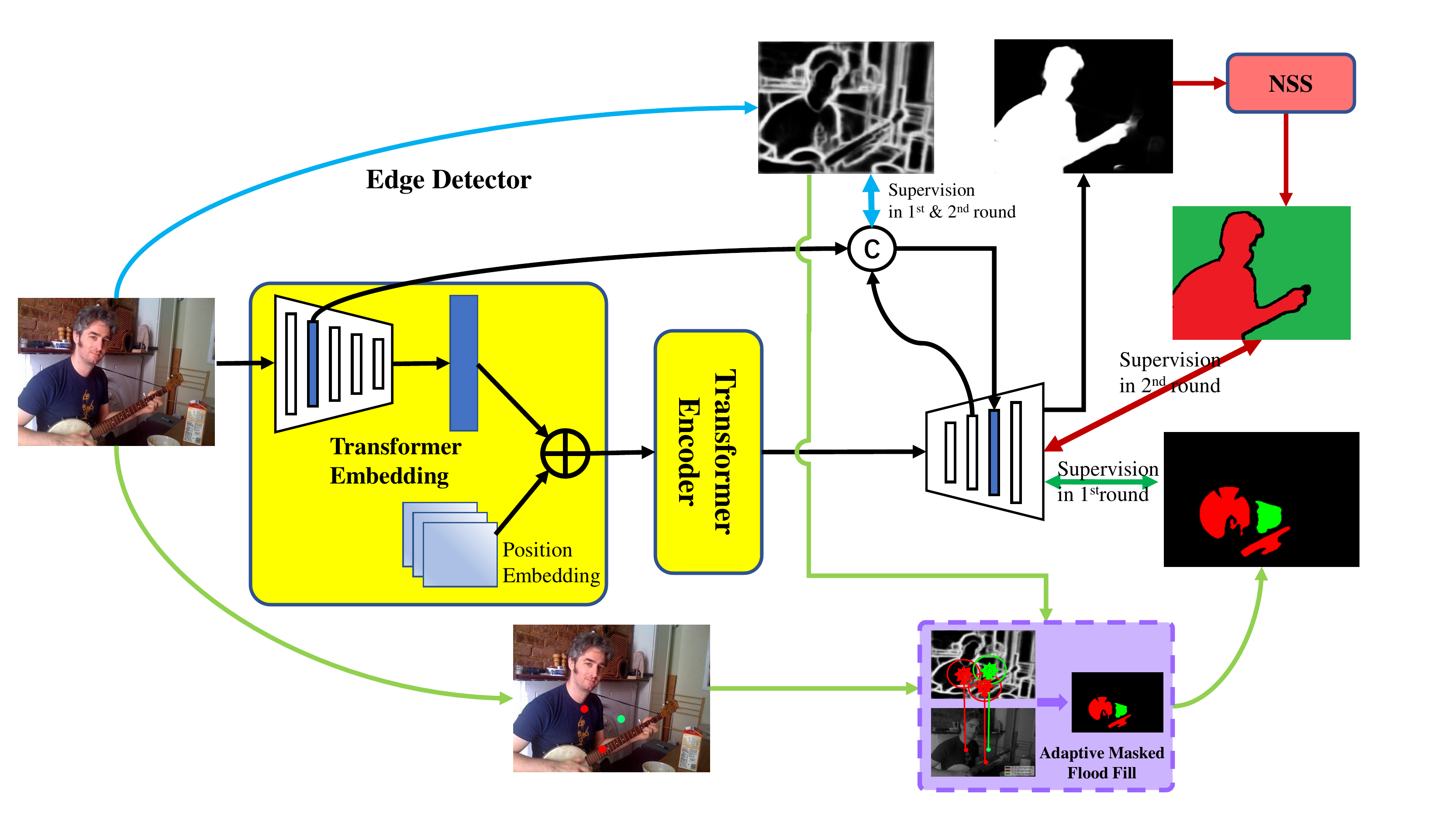} 
		\label{method}
		\caption{The framework of our network and learning strategy. The thick black stream indicates the forward propagation of data in our model. The green stream indicates the parts used in the 1st training round. The blue stream is used in both 1st and 2nd rounds. The red stream is employed in the 2nd training round. The two-way arrows refer to supervision. The yellow box represents the transformer part, in which ResNet-50 is used as the embedding component. The purple box represents the adaptive flood filling. The red box in the upper right corner indicates the Non-Salient object Suppression.}
		
	\end{figure*}

	\section{Methodology}

	\subsection{Adaptive Flood Filling}
	Following the commonly used weakly-supervised dense prediction task,
	we first make pseudo-labels and then use them to train the network.
	As sparse labels only cover a small part of the object region which limits the model’s ability to perceive the structure of the object, \cite{wssa} leverages an edge detector \cite{edge_richer} to generate edges to supervise the training process indirectly supplementing structure. Unlike them, we apply the edges directly to the flood filling. The flood filling starts from a starting node to search its neighborhoods (4 or 8) and extracts the nearby nodes connected to it until all the nodes in the closed area have been processed (Algorithm 1). It is to extract several connected points from one area or to separate them from other adjacent areas. However, since the edges generated by the edge detectors are often discontinuous and blurred (top of Fig. 2), applying it directly to flood filling may result in the whole image being filled. So we designed an adaptive mask, a circle whose radius varies according to the image size, to alleviate this problem. Specifically, the radius $r$ is defined as:
	
	\begin{equation} \label{r_I}
	r(I) = min(\frac{h_I}{\gamma} ,\frac{w_I}{\gamma} )
	\end{equation}
	
	\noindent where $I$ is the input image, $r(I)$ refers to the radius of the mask corresponding to the input image $I$. $h_I$ and $w_I$ represent the length and width of the input image, respectively. $\gamma$ is the hyperparameter. 
	
	\begin{algorithm}[h]
		\label{alg:A}
		\caption{Flood Filling Algorithm}
		\textbf{Input:} Seed point $(x,y)$, image $\mathcal{I}$, seted value $\alpha$, $old$ value $I(x,y)$ \\
		\textbf{Output:} Filled mask $\mathcal{M}$
		
		\begin{algorithmic}[1]
			\STATE   \textit{\textbf{flood filling algorithm}} (($x$,$y$), $\mathcal{I}$, $\alpha$)
			\STATE   \quad \textbf{if} $x \geq 0$ and $x<width$ and $y\geq0$ and $y<height$ 
			\STATE   \quad and $a<\mathcal{I}(x,y)-old<b$ and $\mathcal{I}(x,y)\neq \alpha$ \textbf{then}
			
			\STATE   \quad \quad $\mathcal{M}$(x,y) $\Leftarrow$ $\alpha$
			\STATE   \quad \quad \textit{\textbf{flood filling}} ($(x+1,y)$, $\mathcal{I}$, $\alpha$)
			\STATE   \quad \quad \textit{\textbf{flood filling}} ($(x-1,y)$, $\mathcal{I}$, $\alpha$)
			\STATE   \quad \quad \textit{\textbf{flood filling}} ($(x,y+1)$, $\mathcal{I}$, $\alpha$)
			\STATE   \quad \quad \textit{\textbf{flood filling}} ($(x,y-1)$, $\mathcal{I}$, $\alpha$)
			\STATE  \quad \textbf{end if}
		\end{algorithmic}
	\end{algorithm}
	
	The labeled ground truth of image $I$ can be represented as: $S=\{S_b,S_f^i|i=1,\cdots,N\}$, where $S_b$ and $S_f^i$ refer to the position coordinates of the background pixel and the $i$th labeled salient object, respectively. Then the set of these circle mask can be defined as $\mathcal{M}_S^{r(I)} = C_{S_f^1}^{r(I)} \cup \cdots \cup C_{S_f^N}^{r(I)} \cup C_{S_b}^{r(I)}$, where  $C$ represents the circle that use the lower corner as the center and the upper corner as the radius. Similar to \cite{wssa}, we also use the edge detector \cite{edge_richer} to detect the edges of the image: $e=E(I)$, where $E(\cdot)$ represents the edge detector, $I$ represents the input image, and $e$ represents the detected edges. We use the union of $e$ and $\mathcal{M}_S^{r(I)}$, $E(I) \cup \mathcal{M}_S^{r(I)}$, to divide the image $I$ into multiple connected regions. Then the flood filling is employed to generate the pseudo-label of image $I$  (bottom in Fig. 2):

	\begin{equation} \label{xx}
	g = F(S, E(I) \cup \mathcal{M}_S^{r(I)})
	\end{equation}

	\noindent where $g$ denotes obtained pseudo-label, $F(\cdot)$ denotes the flood filling algorithm, $S$ denotes the labeled ground truth of image $I$.


	\subsection{Transformer-based Point Supervised SOD Model}
	The difficulty of sparse labeling saliency detection lies in the fact that the model can only obtain local ground truth labels and lacks the guidance of global information. We consider that establishing the connection between labeled and unlabeled locations via the similarity between them to obtain the saliency value of the unlabeled region can significantly alleviate this problem. Considering the similarity-based nature of vision transformer (ViT) \cite{vit}, we utilize hyper ViT (i.e., "ResNet-50 + ViT-Base") as our backbone to extract features as well as calculate the self-similarity.
	
	\textbf{Transformer Part.} Specifically, for an input image of size $3\times H\times W$, the CNN embedding part generate $C\times \frac{H}{16}\times \frac{H}{16}$ feature maps. The multi-stage feature of ResNet-50 are denoted as $R=\{R^i|i=1,2,3,4,5\}$. Then, the transformer encoder take as input the summation of position embedding of $C\times \frac{H}{16}\times \frac{H}{16}$ and the flattened features of $C \times\frac{H}{16}\times \frac{H}{16}$ feature maps. 
	After 12 layers of self-attention layers, the transformer encoder part output features of $C \times \frac{H}{16} \times \frac{H}{16}$.

	\textbf{Edge-preserving Decoder.} Edge-preserving decoder consists of two components, a saliency decoder, and an approximate edge detector (see Fig. 2). Saliency decoder is four cascaded convolution layers where each of them followed by batch normalization (BN) layer, ReLU activation, and upsample layer, which takes as input the features of the transformer encoder. And we denote the corresponding features of the saliency decoder at each layer as $D=\{D^i|i=1,2,3,4\}$. 
	
	For the latter part, as the weak annotations lacking structure and details, we develop an edge decoder stream as an approximate edge detector to generate structures by constraining the output using the edges generated by the real edge detector. In detail, the output of approximate edge detector can be represented as $f_e = \sigma(cat(R^3,D^2))$, where $\sigma$ represents a single $3\times3$ convolution layer followed by BN and ReLU layer. The edge map $e$ can be obtained by adding a $3\times3$ convlayer after $f_e$, which is then constrained by the edge map generated from the real edge detector. 
	Then, by merging $f_e$ with $D^3$, $cat(f_e,D^3)$, and passing through the following two convolution layers, the multi-channel feature $f_s$ is obtained. Similarly to $e$, the final saliency map $s$ can be obtained in the same manner.
	

	\subsection{Non-Salient object Suppression (NSS)}
	We observe that due to the sparsity of weakly supervised labels, the supervision signal can only cover a small area of the image, which leads the model only learning to highlight the learned objects while ignoring which objects in the current scene should not be highlighted (red box in Fig. \ref{NSS} (a)). 
	
	To suppress the non-salient object, we propose a simple but effective method by exploiting the location cues provided by the supervision signal and filling the generated highlighted object to suppress the non-salient object. And the obtained salient object regions (red regions in Fig 3. (b)) can be obtained by:

	\begin{equation} \label{pf}
	P_f = F(S-S_b, P^{1st})
	\end{equation}
	where $F(\cdot)$ represents flood filling, $S-S_b = \{S_f^i|i=1,...,N\}$ represents subtraction, $P^{1st}$ represents the saliency maps generated after the first training round and refined by dense CRF \cite{crf}.

	Given that we only provided internal local labels for the salient objects during the first round of training, which may result in the model being unable to distinguish the edges accurately, we perform an expansion operation on $P_f$ with kernel size 10. The expanded regions are designated as the uncertain regions (black regions in Fig. 3 (b)), while the remaining regions are designated as the background region (green regions in Fig. 3 (b)). This is denoted as $P^{2nd}$, which is used as the pseudo-labels for the second training round.

	\begin{figure}[t]
		\centering
		\includegraphics[width=0.4\textwidth]{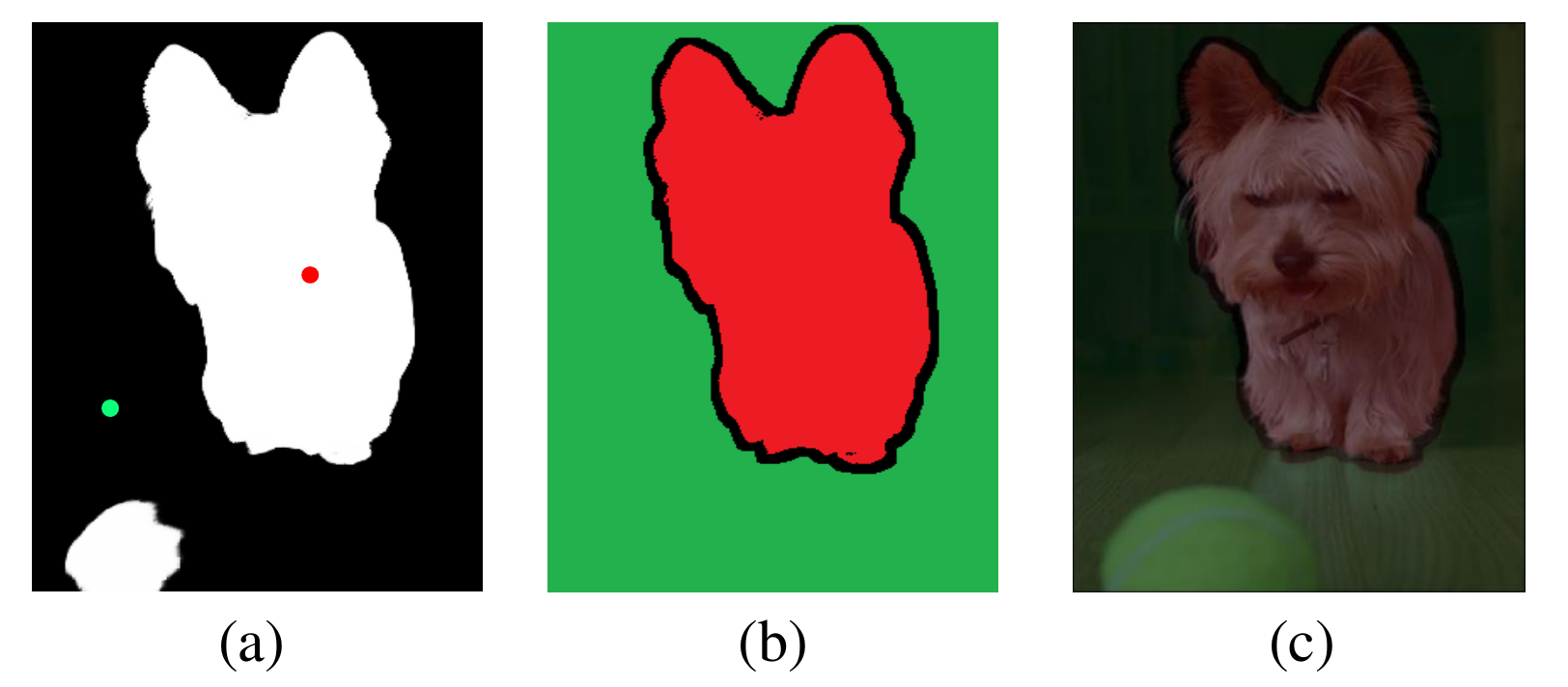} 
		\caption{Illustration of the final pseudo label after NSS. (a) Saliency map generated at the end of first training round. (b) Salient pseudo-label after NSS. (c) The corresponding position of the pseudo label on the original image.}
		\label{nss_viz}
	\end{figure}
	
	As shown in the test examples in Fig. \ref{NSS}, due to the sparseness of the labels, the model tends to detect the non-salient objects. Indeed, the model degenerates into a model that detects the previously learned objects. By reusing the position cues from the supervision points, we can successfully suppress the most non-salient objects utilizing NSS.

	\subsection{Loss Function}
	In our network, binary cross entropy loss, partial cross entropy loss \cite{partial_ce}, and gated CRF loss \cite{scwssod,gate_crf} are employed. For the Edge-preserving decoder stream, we use binary cross entropy loss to constrain $e$:
	
	\begin{equation} \label{partial_ce}
	\mathcal{L}_{bce}=-\sum_{(r,c)}[ylog(e)+(1-y)log(1-e)]
	\end{equation}     
	
	\noindent where $y$ refers to the ground-truth, $e$ represents the predicted edge map, $r$ and $c$ represent the row and column coordinates. For the saliency decoder stream, partial cross entropy loss and gated CRF loss is employed. Partial binary cross-entropy loss is used to focus only on the definite area, while ignoring the uncertain area:
	\begin{equation} \label{partial_ce}
	\mathcal{L}_{pbce}=-\sum_{j\in J}[g_jlog(s_j)+(1-g_j)log(1-s_j)]
	\end{equation}     
	
	\noindent where $J$ represents the labeled area, $g$ refers to the ground truth, $s$ represents the predicted saliency map.

	To obtain better object structure and edges, follow \cite{scwssod}, gated CRF is used in our loss function:
	
	\begin{equation} \label{xx}
	\mathcal{L}_{gcrf} =  \sum_{i}\sum_{j\in K_i} d(i,j)f(i,j)
	\end{equation}
	
	\noindent where $K_i$ denotes the areas covered by $k \times k$ kernel around pixel $i$, $d(i,j)$ is defined as: 
	
	\begin{equation} \label{xx}
	d(i,j) = |s_i - s_j|
	\end{equation}
	
	\noindent where $s_i$ and $s_j$ are the saliency values of $s$ at position i and j, $\left| \cdot \right|$  denotes L1 distance. And $f(i,j)$ refers to the Gaussian kernel bandwidth filter:

	\begin{equation} \label{xx}
	\begin{aligned}
	&f(i,j)= \\
	&\frac{1}{w}\exp(-\frac{\parallel PT(i)-PT(j) \parallel_{2}}{2\sigma_{PT}^2}-\frac{\parallel I(i)-I(j) \parallel_{2}}{2\sigma_I^2})
	\end{aligned}
	\end{equation}

	\noindent where $\frac{1}{w}$ is the weight for normalization, $I(\cdot)$ and $PT(\cdot)$ are the RGB value and position of pixel, $\sigma_{PT}$ and  $\sigma_I$\ are the hyper parameters to control the scale of Gaussian kernels. So the total loss function can be defined as:

	\begin{equation} \label{xx}
	\mathcal{L}_{final}= \alpha_1\mathcal{L}_{bce} + \alpha_2\mathcal{L}_{pbce} + \alpha_3\mathcal{L}_{gcrf}
	\end{equation}
	
	\noindent where $\alpha_1$, $\alpha_2$, $\alpha_3$ are the weights. In our experiments, they are all set to 1.

	\begin{figure*}[htbp]
		\centering
		\includegraphics[width=0.99\textwidth]{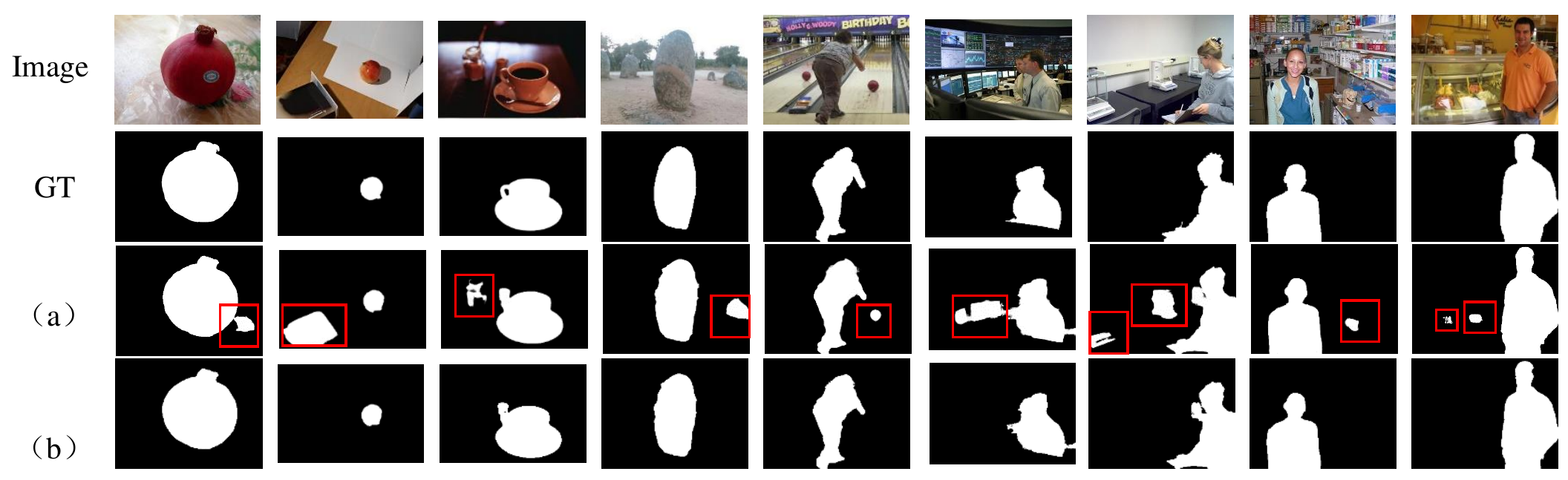} 
		\caption{Illustration of the effect of Non-Salient object Suppression. (a) Pseudo-labels generated at the end of the first round training with point supervision. (b) Salient pseudo-label obtained after passing the pseudo-label of line (a) through Non-Salient object Suppression.}
		\label{NSS}
		\vspace{-0.5cm}
	\end{figure*}

	\section{Experiments}
	
	\subsection{Point-supervised Dataset}
	
	To minimize the labeling time consumption while providing location information of salient objects, we build a Point-supervised Dataset (P-DUTS) by relabeling DUTS \cite{duts} dataset, a widely used saliency detection dataset containing 10553 training images. Four annotators participate in the annotation task, and for each image, we randomly select one of the four annotations to reduce personal bias. In Fig. \ref{annotation}, we show an example of point annotation and compare it with other stronger annotation methods. For each salient object, we only randomly select one-pixel location for labeling (for clarity, we exaggerate the size of the labeled position). Due to the simplicity of our method, even novice annotators can complete an image in 1$\sim $2 seconds on average.

	\subsection{Implementation Details} 
	The proposed model is implemented on the Pytorch toolbox, and our proposed P-DUTS dataset is used as the training set. We train on four TITAN Xp GPUs. For the transformer part, a hyper version transformer ("ResNet50 + ViT-Base") is used by us, and no adjustments are made. We initialize the embedding layers and transformer encoder layers by leveraging the pretrained weight provided by ViT \cite{vit}, which is pre-trained on ImageNet 21K. The maximum learning rate is set to $2.5\times10^{-4}$ for the transformer part and $2.5\times10^{-3}$ for other parts. Warm-up and linear decay strategies are used to adjust the learning rate. Stochastic gradient descent (SGD) is used to train the network, and the following hyper-parameters are used: momentum=0.9, weight decay=$5\times10^{-4}$. Horizontal flip and random crop are used as data augmentation. The batch size is set to 28 and it takes 20 epochs for the first training procedure. The hyperparameter $ \gamma $ of Eq. \ref{r_I} is set to be 5. The second round of training uses the same parameters, but the masks are replaced with refined ones. During testing, we resized each image to 352 $\times$ 352 and then feed it to our network to predict saliency maps. 

	\subsection{Dataset and Evaluation Criteria}
	\textbf{Dataset.} To evaluate the performance, we experiment on five public used benchmark datasets: ECSSD \cite{ecssd}, PASCAL-S \cite{pascal-s}, DUT-O \cite{dut-o}, HKU-IS \cite{hku-is}, and DUTS-test.
	
	\noindent \textbf{Evaluation Criteria.} In our experiments, we compared our model with five state-of-the-art weakly supervised or unsupervised SOD methods (i.e., SCWSSOD \cite{scwssod}, WSSA \cite{wssa}, MFNet \cite{mfnet}, MSW \cite{msw}, SBF \cite{sbf}) and eight fully supervised SOD methods (i.e., GateNet \cite{gatenet}, BASNet \cite{basnet}, AFNet \cite{afnet}, MLMS \cite{mlms}, PiCANet-R \cite{picanet}, DGRL \cite{dgrl}, R$^3$Net \cite{r3+}, RAS \cite{ras}). Five metrics are used: precision-recall curve, maximum F-measure ($F_{\beta}^{max}$), mean F-measure ($mF_\beta$), mean absolute error ($MAE$) and structure-based metric ($S_{m}$) \cite{s-measure}.

	\subsection{Compare with State-of-the-arts}
	
\begin{figure*}[htbp]
		\centering
		\includegraphics[width=0.99\textwidth]{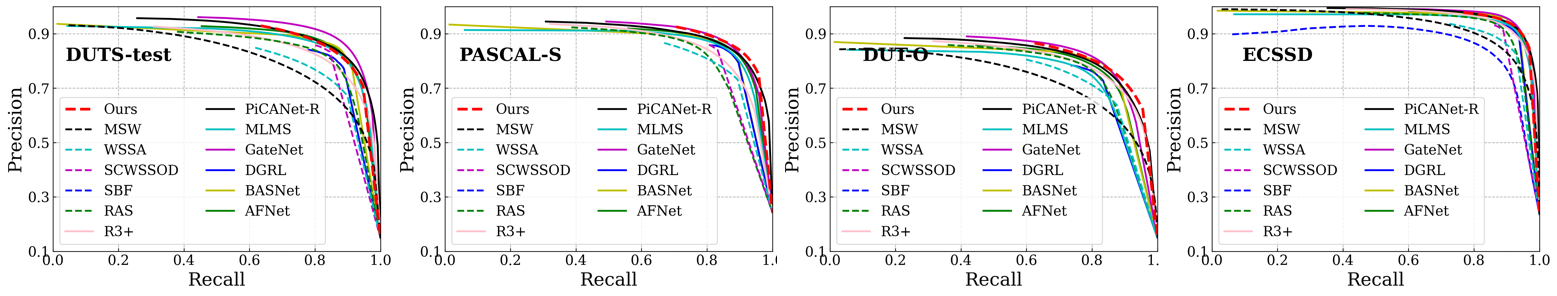} 
		\caption{Illustration of precision-recall curves on the four largest dataset.}
		\label{pr}
	\end{figure*}

	\textbf{Quantitative Comparison.} We compare our method with other state-of-the-art models as shown in Tab. \ref{comparison_SOTA} and Fig. \ref{pr}. For a fair comparison, the saliency results of different compared methods are provided by authors or obtained by running the codes released by authors. The best results are bolded. As can be seen in Tab \ref{comparison_SOTA}, our method outperforms state-of-the-art weakly supervised methods by a large margin. Our method outperforms the previous best model \cite{scwssod} by  2.68\% for $F_\beta^{max}$,  2.75\% for $S_m$, 0.94\% for $mF_\beta$, 0.62 \% for MAE on average of 5 compared dataset. What's more, our method achieves performance on par with recent state-of-the-art model GateNet \cite{gatenet} and even outperforms fully supervised methods in several metrics on multiple data sets as shown in Tab \ref{comparison_SOTA}. And as shown in Fig. \ref{pr}, for quantitative comparison, our method outperforms previous state-of-the-art methods by a large margin.

	\begin{table*}[htbp]
        \centering
        \setlength{\tabcolsep}{1pt}
        \small
            \begin{tabular}{cc |cccccccc| cccccc}
                \hline
                \multirow{3}{*}{Metric} & &\multicolumn{8}{|c|}{Fully Sup.Methods} &\multicolumn{6}{c}{Weakly Sup./Unsup. Methods} \\
                & &RAS &R$^{3}$Net &DGRL &PiCANet &MLMS &AFNet &BASNet &GateNet &SBF &MWS &MFNet &WSSA &SCWSSOD &Ours \\
                & &2018 &2018 &2018 &2018 &2019 &2019 &2019 &2020 &2017 &2019 &2021 &2020 &2021  \\
 
                \hline
 
                \multirow{4}{*}{ECSSD}  &$F_\beta^{max}$ \hfill$\uparrow$   &0.9211 &0.9247 &0.9223 &0.9349 &0.9284 &0.9350 &0.9425 &\textbf{0.9454}      &0.8523 &0.8777 &0.8796   &0.8880 &0.9143  &\textbf{0.9359} \\
                &$mF_\beta$ \hfill$\uparrow$                                &0.9005 &0.9027 &0.9128 &0.9019 &0.9027 &0.9153 &\textbf{0.9306} &0.9251      &0.8220 &0.8060 &0.8603   &0.8801 &0.9091  &\textbf{0.9253} \\
                &$S_m$ \hfill$\uparrow$                                     &0.8928 &0.9030 &0.9026 &0.9170 &0.9111 &0.9134 &0.9162 &\textbf{0.9198}      &0.8323 &0.8275 &0.8345   &0.8655 &0.8818  &\textbf{0.9135} \\
                &MAE \hfill$\downarrow$                                     &0.0564 &0.0556 &0.0408 &0.0464 &0.0445 &0.0418 &\textbf{0.0370} &0.0401      &0.0880 &0.0964 &0.0843   &0.0590 &0.0489  &\textbf{0.0358} \\
                \hline
                \multirow{4}{*}{DUT-O}   &$F_\beta^{max}$\hfill$\uparrow$ &0.7863 &0.7882 &0.7742 &0.8029 &0.7740 &0.7972 &0.8053 &\textbf{0.8181}    &0.6849 &0.7176 &0.7062   &0.7532 &0.7825  &\textbf{0.8086} \\
                &$mF_\beta$\hfill$\uparrow$                                 &0.7621 &0.7533 &0.7658 &0.7628 &0.7470 &0.7763 &\textbf{0.7934} &0.7915      &0.6470 &0.6452 &0.6848   &0.7370 &0.7779  &\textbf{0.7830} \\
                &$S_m$\hfill$\uparrow$                                      &0.8141 &0.8182 &0.8058 &0.8319 &0.8093 &0.8263 &0.8362 &\textbf{0.8381}      &0.7473 &0.7558 &0.7418   &0.7848 &0.8119  &\textbf{0.8243} \\
                &MAE\hfill$\downarrow$                                      &0.0617 &0.0711 &0.0618 &0.0653 &0.0636 &0.0574 &0.0565 &\textbf{0.0549}      &0.1076 &0.1087 &0.0867   &0.0684 &\textbf{0.0602}  &0.0643 \\
                \hline
                \multirow{4}{*}{PASCAL-S}  &$F_\beta^{max}$\hfill$\uparrow$ &0.8291 &0.8374 &0.8486 &0.8573 &0.8552 &0.8629 &0.8539 &\textbf{0.8690}      &0.7593 &0.7840 &0.8046   &0.8088 &0.8406  &\textbf{0.8663} \\
                &$mF_\beta$\hfill$\uparrow$                                 &0.8125 &0.8155 &0.8353 &0.8226 &0.8272 &0.8405 &0.8374 &\textbf{0.8459}      &0.7310 &0.7149 &0.7867   &0.7952 &0.8350  &\textbf{0.8495} \\
                &$S_m$\hfill$\uparrow$                                      &0.7990 &0.8106 &0.8359 &0.8539 &0.8443 &0.8494 &0.8380 &\textbf{0.8580}      &0.7579 &0.7675 &0.7648   &0.7974 &0.8198  &\textbf{0.8529} \\
                &MAE\hfill$\downarrow$                                      &0.1013 &0.1026 &0.0721 &0.0756 &0.0736 &0.0700 &0.0758 &\textbf{0.0674}      &0.1309 &0.1330 &0.1189   &0.0924 &0.0775  &\textbf{0.0647} \\
                \hline
                \multirow{4}{*}{HKU-IS} &$F_\beta^{max}$\hfill$\uparrow$    &0.9128 &0.9096 &0.9103 &0.9185 &0.9207 &0.9226 &0.9284 &\textbf{0.9335}      &-      &0.8560 &0.8766   &0.8805 &0.9084  &\textbf{0.9234} \\
                &$mF_\beta$\hfill$\uparrow$                                 &0.8875 &0.8807 &0.8997 &0.8805 &0.8910 &0.8994 &\textbf{0.9144} &0.9098      &-      &0.7764 &0.8535   &0.8705 &0.9030 &\textbf{0.9131} \\
                &$S_m$\hfill$\uparrow$                                      &0.8874 &0.8920 &0.8945 &0.9044 &0.9065 &0.9053 &0.9089 &\textbf{0.9153}      &-      &0.8182 &0.8465   &0.8649 &0.8820  &\textbf{0.9019} \\
                &MAE\hfill$\downarrow$                                      &0.0454 &0.0478 &0.0356 &0.0433 &0.0387 &0.0358 &\textbf{0.0322} &0.0331      &-      &0.0843 &0.0585   &0.0470 &0.0375  &\textbf{0.0322} \\
                \hline
                \multirow{4}{*}{DUTS-TE} &$F_\beta^{max}$\hfill$\uparrow$   &0.8311 &0.8243 &0.8279 &0.8597 &0.8515 &0.8628 &0.8594 &\textbf{0.8876}      &-      &0.7673 &0.7700   &0.7886 &0.8440  &\textbf{0.8580} \\
                &$mF_\beta$\hfill$\uparrow$                                 &0.8022 &0.7872 &0.8179 &0.8147 &0.8160 &0.8340 &0.8450 &\textbf{0.8558}      &-      &0.7118 &0.7460   &0.7723 &0.8392  &\textbf{0.8402} \\
                &$S_m$\hfill$\uparrow$                                      &0.8385 &0.8360 &0.8417 &0.8686 &0.8617 &0.8670 &0.8656 &\textbf{0.8851}      &-      &0.7587 &0.7747   &0.8034 &0.8405  &\textbf{0.8532} \\
                &MAE\hfill$\downarrow$                                      &0.0594 &0.0664 &0.0497 &0.0506 &0.0490 &0.0458 &0.0476 &\textbf{0.0401}      &-      &0.0912 &0.0765   &0.0622 &0.0487  &\textbf{0.0449} \\
                \hline
        \end{tabular}
        \caption{Quantitative comparison with 13 state-of-the-art methods on ECSSD, DUT-OMRON, PASCAL-S, HKU-IS and DUTS-test. Top results are shown in bold. }
        \label{comparison_SOTA}
    \end{table*}

	\begin{figure*}[h]
		\centering
		\includegraphics[width=0.705\textwidth]{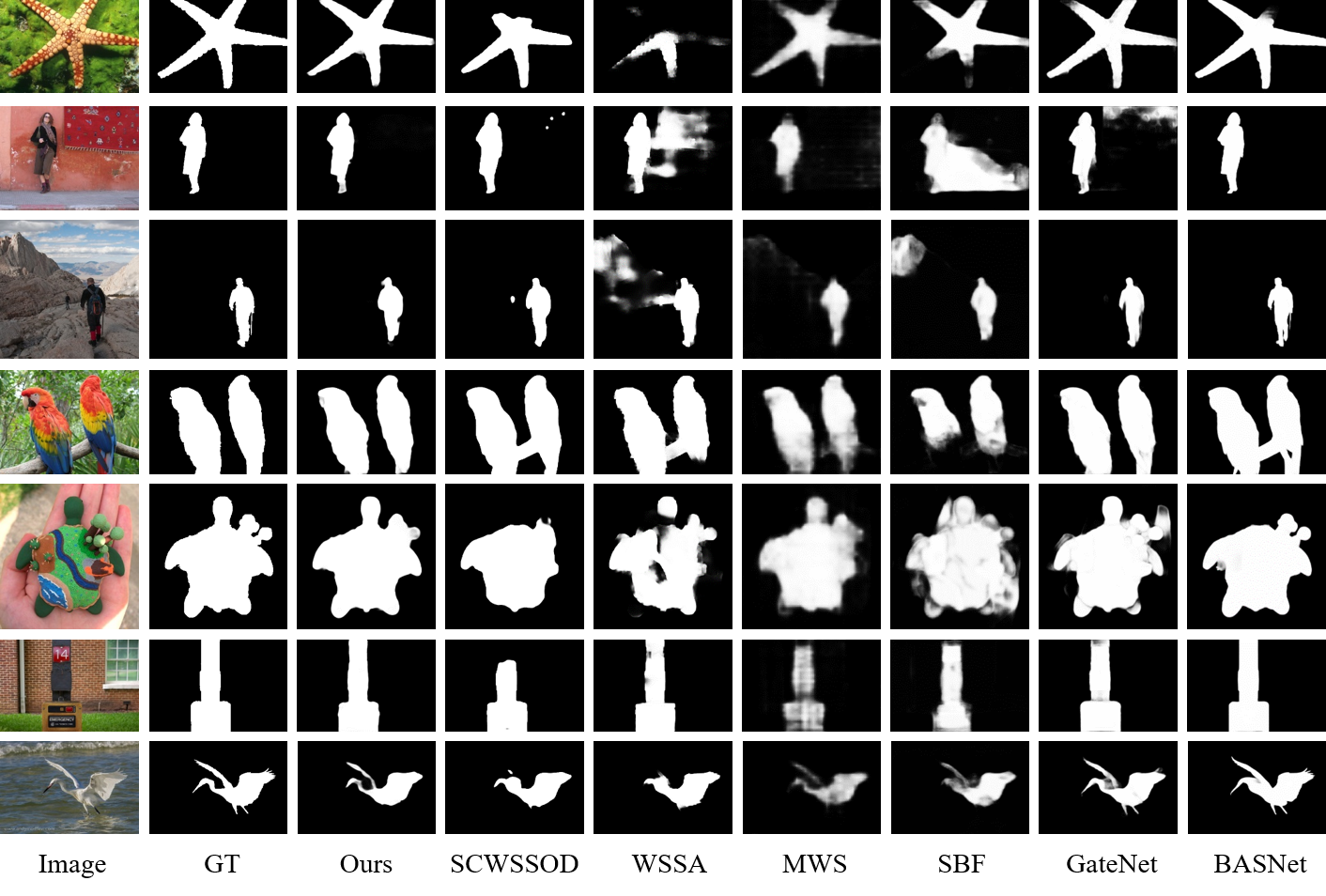} 
		\caption{ Qualitative comparison with different methods.}
		\label{visual_compare}
	\end{figure*}

	\subsubsection{Qualitative Evaluation.}
	As shown in Fig. \ref{visual_compare}, our method obtains more accurate and complete saliency maps compared with other state-of-the-art weakly supervised or unsupervised methods and even surpasses recently fully supervised state-of-the-art methods. Row 1, 5, and 6 demonstrate the ability of our model to capture the overall salient object, and even objects (row 5) with extremely complex textures can be accurately segmented by our method. Further, thanks to our NSS method, our method can precisely extract salient objects and suppress non-salient objects (rows 2-4). The last row shows the ability of our model to extract the details.

	\subsection{Ablation Study}
	
	\subsubsection{Effectiveness of Edge-preserving Decoder.}
	Note that, we use the results of the first round to evaluate the performance because if the first round produces bad results it will directly affect the results of the second round. In Tab. \ref{edge_decoder}, we evaluate the influence of the Edge-preserving decoder stream. The supervision of the edge generated by the edge detector is removed. At this point, the edge-preserving decoder can only aggregate shallow features and cannot explicitly detect edge features, and the overall performance decreases.

	\subsubsection{Effectiveness of NSS.}
	Since NSS is used in the 2nd training round, we perform ablation experiments based on the results generated in the 1st round. The ablation study results are listed in Tab. \ref{nss}. Line 1 indicates we directly use the pseudo-labels generated by the 1st round without CRF processing ("E" means using edge constraints). Line 2 indicates we use pseudo-labels refined by dense CRF. Line 3 represents the standard method in this paper. The last line represents we directly employ $P_f$ (Eq. \ref{pf}) as ground truth with no uncertain areas. 
	The comparison of line 1 and line 2 shows that using CRF to optimize the pseudo labels can optimize the second training round.
	The 3rd row works better than the first two rows, which shows the validity of NSS. We also illustrate the effectiveness of NSS in Fig. \ref{NSS}. 

		
				
				
		

	\begin{table}
	\centering
	\small
		
		\label{tab:freq}
        \setlength{\tabcolsep}{1mm}{
			\begin{tabular}{c|ccc| ccc}
				\hline
				\multirow{2}{*}{Model}    &\multicolumn{3}{c|}{PASCAL-S}   &\multicolumn{3}{c}{DUT-OMRON}\cr
				&$F_\beta^{max} \uparrow$   &$S_m \uparrow$ &MAE$\downarrow$  &$F_\beta^{max} \uparrow$  &$S_m \uparrow$ &MAE $\downarrow$\\
				\hline 
				
				w/ &0.8636	&0.8498	&0.0671 &0.8034 &0.8150 &0.0717 \\ 
				w/o &0.8629	&0.8454	&0.0703 &0.8049 &0.8054 &0.0769 \\
				
				\hline
		\end{tabular}}
		
		\caption{ The effectiveness Edge-preserving Decoder.}
		\label{edge_decoder}
	\end{table}

		
				

	\begin{table}
	\centering
	\small
		\label{tab:freq}

		
		\setlength{\tabcolsep}{0.5mm}{
			\begin{tabular}{c|ccc| ccc}
				\hline
				\multirow{2}{*}{Model}    &\multicolumn{3}{c|}{DUTS-test}   &\multicolumn{3}{c}{ECSSD}\cr
				&$F_\beta^{max} \uparrow$   &$S_m \uparrow$ &MAE$\downarrow$  &$F_\beta^{max} \uparrow$ &$S_m \uparrow$ &MAE $\downarrow$\\
				\hline
				+E        &0.8588  &0.8559 &0.0502 &0.9342  &0.9134 &0.0407\\
				+E+CRF    &0.8606  &0.8510 &0.0463 &0.935  &0.9131 &0.0385 \\
				+E+CRF+NSS$^*$  &0.8580  &0.8532 &0.0449 &0.9359 &0.9135 &0.0358 \\
				\hline
				
				\makecell[c]{+E+CRF+NSS} &0.8586	&0.8541	&0.0457 &0.9365	&0.9121	&0.0382 \\
				\hline
		\end{tabular}}
		\caption{Ablation study for NSS.} 
		\label{nss}
	\end{table}

	\begin{table}
		\centering
		\small
		
		\label{tab:freq}
        \setlength{\tabcolsep}{1mm}{
			\begin{tabular}{c|ccc| ccc}
				\hline
				\multirow{2}{*}{$\gamma$}    &\multicolumn{3}{c|}{DUTS-test}   &\multicolumn{3}{c}{ECSSD}\cr
				&$F_\beta^{max} \uparrow$   &$S_m \uparrow$ &MAE$\downarrow$  &$F_\beta^{max} \uparrow$  &$S_m \uparrow$ &MAE $\downarrow$\\
				\hline
				3   &0.8467	&0.8421	&0.0559 &0.9299		&0.9085	&0.0415\\
				4  &0.8497	&0.8439	&0.0543 &0.9304		&0.9088	&0.0415\\
				5 &0.8530 &0.8516 &0.0488 &0.9326 &0.9095 &0.0384\\
				6 &0.8480 &0.8416 &0.0558 &0.9293  &0.9079 &0.0421\\
				\hline
		\end{tabular}}
		
		\caption{ The impact of $\gamma$.}
		\label{gama}
	\end{table}

	\subsubsection{Impact of the Hyperparameter $\gamma$.}
	We test the effect of different values of $\gamma$ on the first round of training in Tab. \ref{gama}. Too small $\gamma$ will generate small pseudo labels providing too little supervision information, and too large $\gamma$ will make the pseudo labels contain the wrong areas. Both of these cases affect the model's performance.

	\section{Conclusions}
	
	We propose a point-supervised framework to detect salient objects in this paper. An adaptive flood filling algorithm and a transform-based point-supervised model are designed for pseudo-label generation and saliency detection. To uncover the problem of degradation of the existing weakly supervised SOD models, we propose the Non-Salient object Suppression (NSS) technique to explicitly filter out the non-salient but detected objects. We build a new point-supervised saliency dataset P-DUTS for model training. The proposed method not only outperforms existing state-of-the-art weakly-supervised methods with stronger supervision.
	
	\section{Acknowledgements}
    This work was supported by the National Key R\&D Program of China (2020AAA0108301), the National Natural Science Foundation of China (No.62072112), Scientific and technological innovation action plan of  Shanghai Science and Technology Committee (No.205111031020).

	\bibliography{ours_aaai22}

\begin{thebibliography}{43}
\providecommand{\natexlab}[1]{#1}

\bibitem[{Bearman et~al.(2016)Bearman, Russakovsky, Ferrari, and
  Fei-Fei}]{what_point}
Bearman, A.; Russakovsky, O.; Ferrari, V.; and Fei-Fei, L. 2016.
\newblock What’s the point: Semantic segmentation with point supervision.
\newblock In \emph{European conference on computer vision}, 549--565. Springer.

\bibitem[{Benenson, Popov, and Ferrari(2019)}]{point_large}
Benenson, R.; Popov, S.; and Ferrari, V. 2019.
\newblock Large-scale interactive object segmentation with human annotators.
\newblock In \emph{Proceedings of the IEEE/CVF Conference on Computer Vision
  and Pattern Recognition}, 11700--11709.

\bibitem[{Carion et~al.(2020)Carion, Massa, Synnaeve, Usunier, Kirillov, and
  Zagoruyko}]{detr}
Carion, N.; Massa, F.; Synnaeve, G.; Usunier, N.; Kirillov, A.; and Zagoruyko,
  S. 2020.
\newblock End-to-end object detection with transformers.
\newblock In \emph{European Conference on Computer Vision}, 213--229. Springer.

\bibitem[{Chen et~al.(2018)Chen, Tan, Wang, and Hu}]{ras}
Chen, S.; Tan, X.; Wang, B.; and Hu, X. 2018.
\newblock Reverse attention for salient object detection.
\newblock In \emph{European Conference on Computer Vision}, 234--250.

\bibitem[{Deng et~al.(2018)Deng, Hu, Zhu, Xu, Qin, Han, and Heng}]{r3+}
Deng, Z.; Hu, X.; Zhu, L.; Xu, X.; Qin, J.; Han, G.; and Heng, P.-A. 2018.
\newblock R3net: Recurrent residual refinement network for saliency detection.
\newblock In \emph{Proceedings of the 27th International Joint Conference on
  Artificial Intelligence}, 684--690.

\bibitem[{Dosovitskiy et~al.(2020)Dosovitskiy, Beyer, Kolesnikov, Weissenborn,
  and Houlsby}]{vit}
Dosovitskiy, A.; Beyer, L.; Kolesnikov, A.; Weissenborn, D.; and Houlsby, N.
  2020.
\newblock An Image is Worth 16x16 Words: Transformers for Image Recognition at
  Scale.
\newblock \emph{International Conference on Learning Representations}.

\bibitem[{Fan et~al.(2017)Fan, Cheng, Liu, Li, and Borji}]{s-measure}
Fan, D.-P.; Cheng, M.-M.; Liu, Y.; Li, T.; and Borji, A. 2017.
\newblock Structure-measure: A new way to evaluate foreground maps.
\newblock In \emph{Proceedings of the IEEE international conference on computer
  vision}, 4548--4557.

\bibitem[{Feng, Lu, and Ding(2019)}]{afnet}
Feng, M.; Lu, H.; and Ding, E. 2019.
\newblock Attentive feedback network for boundary-aware salient object
  detection.
\newblock In \emph{Proceedings of the IEEE/CVF Conference on Computer Vision
  and Pattern Recognition}, 1623--1632.

\bibitem[{Flores et~al.(2019)Flores, Gonzalezgarcia, De~Weijer, and
  Raducanu}]{saliency-recognition}
Flores, C.~F.; Gonzalezgarcia, A.; De~Weijer, J.~V.; and Raducanu, B. 2019.
\newblock Saliency for fine-grained object recognition in domains with scarce
  training data.
\newblock \emph{Pattern Recognition}, 62--73.

\bibitem[{He and Liu(2020)}]{person-re-id}
He, L.; and Liu, W. 2020.
\newblock Guided saliency feature learning for person re-identification in
  crowded scenes.
\newblock In \emph{European Conference on Computer Vision}, 357--373.

\bibitem[{Itti, Koch, and Niebur(1998)}]{itti}
Itti, L.; Koch, C.; and Niebur, E. 1998.
\newblock A model of saliency-based visual attention for rapid scene analysis.
\newblock \emph{IEEE Transactions on pattern analysis and machine
  intelligence}, 20(11): 1254--1259.

\bibitem[{Kr{\"a}henb{\"u}hl and Koltun(2011)}]{crf}
Kr{\"a}henb{\"u}hl, P.; and Koltun, V. 2011.
\newblock Efficient inference in fully connected crfs with gaussian edge
  potentials.
\newblock In \emph{Advances in Neural Information Processing Systems},
  109--117.

\bibitem[{Li, Xie, and Lin(2018)}]{wsi}
Li, G.; Xie, Y.; and Lin, L. 2018.
\newblock Weakly supervised salient object detection using image labels.
\newblock In \emph{Thirty-second AAAI conference on artificial intelligence}.

\bibitem[{Li and Yu(2015)}]{hku-is}
Li, G.; and Yu, Y. 2015.
\newblock Visual saliency based on multiscale deep features.
\newblock In \emph{Proceedings of the IEEE/CVF Conference on Computer Vision
  and Pattern Recognition}, 5455--5463.

\bibitem[{Li et~al.(2014)Li, Hou, Koch, Rehg, and Yuille}]{pascal-s}
Li, Y.; Hou, X.; Koch, C.; Rehg, J.~M.; and Yuille, A.~L. 2014.
\newblock The secrets of salient object segmentation.
\newblock In \emph{Proceedings of the IEEE/CVF Conference on Computer Vision
  and Pattern Recognition}, 280--287.

\bibitem[{Li, Chen, and Koltun(2018)}]{point_latent}
Li, Z.; Chen, Q.; and Koltun, V. 2018.
\newblock Interactive image segmentation with latent diversity.
\newblock In \emph{Proceedings of the IEEE Conference on Computer Vision and
  Pattern Recognition}, 577--585.

\bibitem[{Liew et~al.(2017)Liew, Wei, Xiong, Ong, and Feng}]{point_regional}
Liew, J.; Wei, Y.; Xiong, W.; Ong, S.-H.; and Feng, J. 2017.
\newblock Regional interactive image segmentation networks.
\newblock In \emph{2017 IEEE international conference on computer vision
  (ICCV)}, 2746--2754. IEEE Computer Society.

\bibitem[{Liu, Han, and Yang(2018)}]{picanet}
Liu, N.; Han, J.; and Yang, M.-H. 2018.
\newblock Picanet: Learning pixel-wise contextual attention for saliency
  detection.
\newblock In \emph{Proceedings of the IEEE/CVF Conference on Computer Vision
  and Pattern Recognition}, 3089--3098.

\bibitem[{Liu et~al.(2017)Liu, Cheng, Hu, Wang, and Bai}]{edge_richer}
Liu, Y.; Cheng, M.-M.; Hu, X.; Wang, K.; and Bai, X. 2017.
\newblock Richer convolutional features for edge detection.
\newblock In \emph{Proceedings of the IEEE conference on computer vision and
  pattern recognition}, 3000--3009.

\bibitem[{Liu et~al.(2021{\natexlab{a}})Liu, Wang, Cao, Liang, and
  Lau}]{tipweaklyboundingbox}
Liu, Y.; Wang, P.; Cao, Y.; Liang, Z.; and Lau, R.~W. 2021{\natexlab{a}}.
\newblock Weakly-Supervised Salient Object Detection With Saliency Bounding
  Boxes.
\newblock \emph{IEEE Transactions on Image Processing}, 30: 4423--4435.

\bibitem[{Liu et~al.(2021{\natexlab{b}})Liu, Lin, Cao, Hu, Wei, Zhang, Lin, and
  Guo}]{swin}
Liu, Z.; Lin, Y.; Cao, Y.; Hu, H.; Wei, Y.; Zhang, Z.; Lin, S.; and Guo, B.
  2021{\natexlab{b}}.
\newblock Swin transformer: Hierarchical vision transformer using shifted
  windows.
\newblock In \emph{Proceedings of the IEEE/CVF International Conference on
  Computer Vision}, 10012--10022.

\bibitem[{Maninis et~al.(2018)Maninis, Caelles, Pont-Tuset, and
  Van~Gool}]{point_extreme}
Maninis, K.-K.; Caelles, S.; Pont-Tuset, J.; and Van~Gool, L. 2018.
\newblock Deep extreme cut: From extreme points to object segmentation.
\newblock In \emph{Proceedings of the IEEE Conference on Computer Vision and
  Pattern Recognition}, 616--625.

\bibitem[{Obukhov et~al.(2019)Obukhov, Georgoulis, Dai, and
  Van~Gool}]{gate_crf}
Obukhov, A.; Georgoulis, S.; Dai, D.; and Van~Gool, L. 2019.
\newblock Gated CRF loss for weakly supervised semantic image segmentation.
\newblock \emph{arXiv preprint arXiv:1906.04651}.

\bibitem[{Piao et~al.(2021)Piao, Wang, Zhang, and Lu}]{mfnet}
Piao, Y.; Wang, J.; Zhang, M.; and Lu, H. 2021.
\newblock MFNet: Multi-filter Directive Network for Weakly Supervised Salient
  Object Detection.
\newblock In \emph{Proceedings of the IEEE/CVF International Conference on
  Computer Vision}, 4136--4145.

\bibitem[{Qian et~al.(2019)Qian, Wei, Shi, Li, Liu, and Huang}]{metric_point}
Qian, R.; Wei, Y.; Shi, H.; Li, J.; Liu, J.; and Huang, T. 2019.
\newblock Weakly supervised scene parsing with point-based distance metric
  learning.
\newblock In \emph{Proceedings of the AAAI Conference on Artificial
  Intelligence}, volume~33, 8843--8850.

\bibitem[{Qin et~al.(2019)Qin, Zhang, Huang, Gao, Dehghan, and
  Jagersand}]{basnet}
Qin, X.; Zhang, Z.; Huang, C.; Gao, C.; Dehghan, M.; and Jagersand, M. 2019.
\newblock BASNet: Boundary-Aware Salient Object Detection.
\newblock In \emph{Proceedings of the IEEE/CVF Conference on Computer Vision
  and Pattern Recognition}, 7479--7489.

\bibitem[{Tang et~al.(2018)Tang, Djelouah, Perazzi, Boykov, and
  Schroers}]{partial_ce}
Tang, M.; Djelouah, A.; Perazzi, F.; Boykov, Y.; and Schroers, C. 2018.
\newblock Normalized cut loss for weakly-supervised cnn segmentation.
\newblock In \emph{Proceedings of the IEEE Conference on Computer Vision and
  Pattern Recognition}, 1818--1827.

\bibitem[{Wang et~al.(2017)Wang, Lu, Wang, Feng, Wang, Yin, and Ruan}]{duts}
Wang, L.; Lu, H.; Wang, Y.; Feng, M.; Wang, D.; Yin, B.; and Ruan, X. 2017.
\newblock Learning to detect salient objects with image-level supervision.
\newblock In \emph{Proceedings of the IEEE/CVF Conference on Computer Vision
  and Pattern Recognition}, 136--145.

\bibitem[{Wang et~al.(2018)Wang, Zhang, Wang, Lu, Yang, Ruan, and Borji}]{dgrl}
Wang, T.; Zhang, L.; Wang, S.; Lu, H.; Yang, G.; Ruan, X.; and Borji, A. 2018.
\newblock Detect globally, refine locally: A novel approach to saliency
  detection.
\newblock In \emph{Proceedings of the IEEE/CVF Conference on Computer Vision
  and Pattern Recognition}, 3127--3135.

\bibitem[{Wang et~al.(2021)Wang, Lai, Fu, Shen, and Ling}]{depth-survey}
Wang, W.; Lai, Q.; Fu, H.; Shen, J.; and Ling, H. 2021.
\newblock Salient Object Detection in the Deep Learning Era: An In-Depth
  Survey.
\newblock \emph{IEEE Transactions on Pattern Analysis and Machine
  Intelligence}.

\bibitem[{Wei, Wang, and Huang(2020)}]{f3net}
Wei, J.; Wang, S.; and Huang, Q. 2020.
\newblock F{\({^3}\)}Net: Fusion, Feedback and Focus for Salient Object
  Detection.
\newblock In \emph{Proceedings of the AAAI Conference on Artificial
  Intelligence}, 12321--12328.

\bibitem[{{Wei} et~al.(2020){Wei}, {Wang}, {Wu}, {Su}, {Huang}, and
  {Tian}}]{ldf}
{Wei}, J.; {Wang}, S.; {Wu}, Z.; {Su}, C.; {Huang}, Q.; and {Tian}, Q. 2020.
\newblock F{\({^3}\)}Net: Fusion, Feedback and Focus for Salient Object
  Detection.
\newblock In \emph{Proceedings of the IEEE/CVF Conference on Computer Vision
  and Pattern Recognition}, 13022--13031.

\bibitem[{Wu et~al.(2019)Wu, Feng, Guan, Wang, Lu, and Ding}]{mlms}
Wu, R.; Feng, M.; Guan, W.; Wang, D.; Lu, H.; and Ding, E. 2019.
\newblock A mutual learning method for salient object detection with
  intertwined multi-supervision.
\newblock In \emph{Proceedings of the IEEE/CVF Conference on Computer Vision
  and Pattern Recognition}, 8150--8159.

\bibitem[{Yan et~al.(2013)Yan, Xu, Shi, and Jia}]{ecssd}
Yan, Q.; Xu, L.; Shi, J.; and Jia, J. 2013.
\newblock Hierarchical saliency detection.
\newblock In \emph{Proceedings of the IEEE/CVF Conference on Computer Vision
  and Pattern Recognition}, 1155--1162.

\bibitem[{Yang et~al.(2013)Yang, Zhang, Lu, Ruan, and Yang}]{dut-o}
Yang, C.; Zhang, L.; Lu, H.; Ruan, X.; and Yang, M.-H. 2013.
\newblock Saliency detection via graph-based manifold ranking.
\newblock In \emph{Proceedings of the IEEE/CVF Conference on Computer Vision
  and Pattern Recognition}, 3166--3173.

\bibitem[{Yu et~al.(2021)Yu, Zhang, Xiao, and Lim}]{scwssod}
Yu, S.; Zhang, B.; Xiao, J.; and Lim, E.~G. 2021.
\newblock Structure-consistent weakly supervised salient object detection with
  local saliency coherence.
\newblock In \emph{Proceedings of the AAAI Conference on Artificial
  Intelligence (AAAI)}.

\bibitem[{Zeng et~al.(2019)Zeng, Zhuge, Lu, Zhang, Qian, and Yu}]{msw}
Zeng, Y.; Zhuge, Y.; Lu, H.; Zhang, L.; Qian, M.; and Yu, Y. 2019.
\newblock Multi-source weak supervision for saliency detection.
\newblock In \emph{Proceedings of the IEEE/CVF Conference on Computer Vision
  and Pattern Recognition}, 6074--6083.

\bibitem[{Zhang, Han, and Zhang(2017)}]{sbf}
Zhang, D.; Han, J.; and Zhang, Y. 2017.
\newblock Supervision by fusion: Towards unsupervised learning of deep salient
  object detector.
\newblock In \emph{Proceedings of the IEEE International Conference on Computer
  Vision}, 4048--4056.

\bibitem[{Zhang et~al.(2020)Zhang, Yu, Li, Song, Liu, and Dai}]{wssa}
Zhang, J.; Yu, X.; Li, A.; Song, P.; Liu, B.; and Dai, Y. 2020.
\newblock Weakly-supervised salient object detection via scribble annotations.
\newblock In \emph{Proceedings of the IEEE/CVF conference on computer vision
  and pattern recognition}, 12546--12555.

\bibitem[{Zhao et~al.(2020)Zhao, Pang, Zhang, Lu, and Zhang}]{gatenet}
Zhao, X.; Pang, Y.; Zhang, L.; Lu, H.; and Zhang, L. 2020.
\newblock Suppress and balance: A simple gated network for salient object
  detection.
\newblock In \emph{European Conference on Computer Vision}, 35--51.

\bibitem[{Zheng et~al.(2021)Zheng, Lu, Zhao, Zhu, Luo, Wang, Fu, Feng, Xiang,
  Torr et~al.}]{setr}
Zheng, S.; Lu, J.; Zhao, H.; Zhu, X.; Luo, Z.; Wang, Y.; Fu, Y.; Feng, J.;
  Xiang, T.; Torr, P.~H.; et~al. 2021.
\newblock Rethinking semantic segmentation from a sequence-to-sequence
  perspective with transformers.
\newblock In \emph{Proceedings of the IEEE/CVF Conference on Computer Vision
  and Pattern Recognition}, 6881--6890.

\bibitem[{Zhou et~al.(2016)Zhou, Khosla, Lapedriza, Oliva, and Torralba}]{cam}
Zhou, B.; Khosla, A.; Lapedriza, A.; Oliva, A.; and Torralba, A. 2016.
\newblock Learning deep features for discriminative localization.
\newblock In \emph{Proceedings of the IEEE conference on computer vision and
  pattern recognition}, 2921--2929.

\bibitem[{Zhou et~al.(2020)Zhou, Zhang, Jiang, Zhang, and Fan}]{re-caption}
Zhou, L.; Zhang, Y.; Jiang, Y.; Zhang, T.; and Fan, W. 2020.
\newblock Re-Caption: Saliency-Enhanced Image Captioning Through Two-Phase
  Learning.
\newblock \emph{IEEE Transactions on Image Processing}, 694--709.

\end{thebibliography}

\end{document}